\documentclass[11pt,a4paper]{article}
\usepackage[hyperref]{naaclhlt2019}
\usepackage{times}
\usepackage{latexsym}

\usepackage{url}

\usepackage{subfigure}
\usepackage{amsmath,amssymb,amsfonts}
\usepackage{algorithmic}
\usepackage{graphicx}
\usepackage{array}

\usepackage{dblfloatfix}    

\aclfinalcopy 
\def\aclpaperid{26} 

\title{Viable Threat on News Reading: Generating Biased News Using Natural Language Models}

\author{ Saurabh Gupta$^1$, Huy H. Nguyen$^{2,4}$, Junichi Yamagishi$^{2,4}$  and Isao Echizen$^{2,3,4}$ \\
$^1$ Indraprastha Institute of Information Technology - Delhi, Delhi, India \\
$^2$ National Institute of Informatics, Tokyo, Japan; $^3$ {University of Tokyo, Japan} \\
$^4$ The Graduate University for Advanced Studies, SOKENDAI, Kanagawa, Japan \\
{\tt saurabhg@iiitd.ac.in},  {\tt \{nhhuy,jyamagis,iechizen\}@nii.ac.jp}}

\begin{document}
\maketitle
\begin{abstract}

Recent advancements in natural language generation has raised serious concerns. High-performance language models are widely used for language generation tasks because they are able to produce fluent and meaningful sentences. These models are already being used to create fake news. They can also be exploited to generate biased news, which can then be used to attack news aggregators to change their reader's behavior and influence their bias.  In this paper, we use a threat model to demonstrate that the publicly available language models can reliably generate biased news content based on an input original news. We also show that a large number of high-quality biased news articles can be generated using controllable text generation. A subjective evaluation with 80 participants demonstrated that the generated biased news is generally fluent, and a bias evaluation with 24 participants demonstrated that the bias (left or right) is usually evident in the generated articles and can be easily identified.

  
\end{abstract}

\section{Introduction}
\label{section:introduction}








Natural language generation is defined as the creation of understandable text using a language model (LM) trained on a large collection of texts. An (LM) is a probability distribution over a sequence of words. Given a set of training text sequences, we can train an LM to produce texts similar to the training data. Researchers have used deep learning algorithms to generate more fluent and semantically meaningful texts than those generated using conventional methods like n-grams \cite{lu2018neural}. Such LMs are being used to generate image captions \cite{vinyals2015show}, perform machine translations \cite{bahdanau2014neural}, paraphrase and summarize text \cite{zhang2017semantic}. High performance LMs can generate fake news, fake reviews, and fake comments \cite{adelani2020generating,zellers2019defending}.

Recent studies have revealed various types of bias in top US news sources, which often report political news in a biased way, for example, attention can be drawn to particular events and entities while ignoring others \cite{ribeiro2018media,groseclose2005measure,kulshrestha2017quantifying}. The selection of what to report about an entity (positive or negative) produces bias. There are two major political sides in the U.S.: \textit{Democrats} on the left and \textit{Republicans} on the right.

The news aggregating platforms like Google News and Yahoo News are the most viewed news websites in U.S. with 150 and 175 million unique visitors every month, respectively \cite{statista}. They offer content relevant to a wide range of global audiences, and therefore, they have a responsibility to maintain the same sentiment and bias.  However, they can utilize language models to generate biased content (news headlines and articles) to model the behavior of their readers. Exposure to biased news is very harmful as it can increase/flip the political bias of a reader \cite{bail2018exposure}. For example, \cite{wong2019cambridge} found that exposure to biased news can alter the political inclinations of people, and  \cite{doi:10.1177/107769909407100109} found that false representation of news from a news source can lead to broken trust between the reader and the news source.

\begin{figure*}[htbp]
    \centering
    \includegraphics[width=0.9\linewidth]{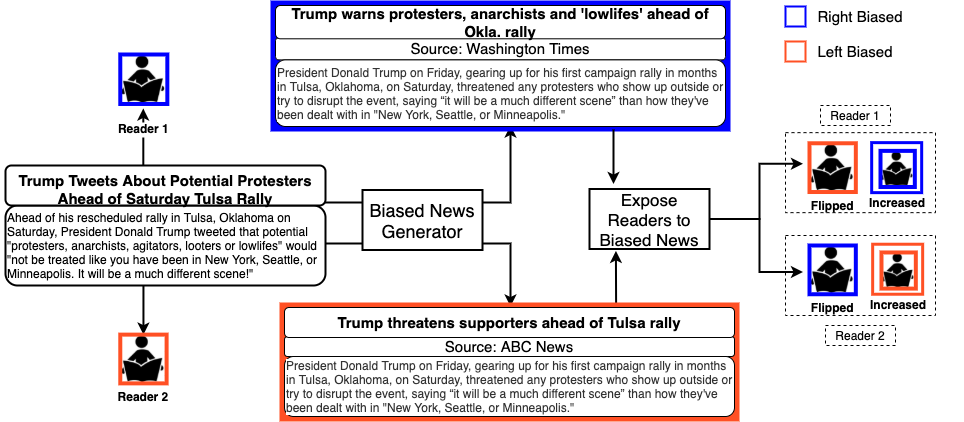}
    \caption{ Proposed threat model. Original news is used as seed by the ``Biased News Generator'' (explained in Section \ref{section:generation}) to generate left or right biased news. Readers are then exposed to the generated biased news to change their original bias (either flip or increase).}
    \label{fig:threat_model}
\end{figure*}

Previous works on media bias mostly focused on detecting bias either by using cues from the social media presence of the news sources \cite{kulshrestha2017quantifying,an2012visualizing,ribeiro2018media}, or by analyzing how bias is manifested within each news article \cite{chen2018learning}. \newcite{chen2018learning} focused on flipping the bias of news headlines, which is a short one line text. \newcite{bail2018exposure} showed that exposure to opposing views can increase political polarization. To the best of our knowledge, ours is the first attempt at generating full length biased news articles using high performance language models. 

\textit{Our Contribution.} In this paper, we use a threat model (Figure \ref{fig:threat_model}) to demonstrate that publicly available language models can reliably generate biased news content based on original news. In an ideal scenario, a user consumes original news from an aggregator and develops a confirmation bias \cite{nickerson1998confirmation} about entities mentioned in the news. If the news complements their bias, they likely jump to the original source to continue reading \cite{swire2017role}. Our threat model, we assume that the attacker is able to access the original news and have control over what a user will see when visiting the aggregator’s platform. In this scenario, the attacker can rework the original news, by either shifting its bias farther than it originally was~\cite{levendusky2013partisan}, or by flipping its original bias~\cite{bail2018exposure}. The attacker is also assumed to be able to access a large collection of news articles labeled with the bias (left or right) to use for training an LM. The attacker uses the original news as input to the LM for using as context to generate biased news. Finally, the attacker exposes readers to the generated biased news.

To generate biased news, we fine-tuned the GPT-2 language model (LM) \cite{radford2019language} to create two different LMs, each trained on a specific type of biased news. We used an API built on a RoBERTa–based model \cite{DBLP:journals/corr/abs-1907-11692} (explained in a later section) to classify the generated news as left or right biased. However, generating only the text for news is not enough. Therefore, we then fine-tuned another generative model, known as GROVER \cite{zellers2019defending}, which enables controllable generation of an entire news article – the body, title, news source, publication date, and author list. Finally, we performed a subjective evaluation with 80 participants - 32 native and 48 non-native English speakers. The results show that the news articles generated by the models (machine-generated news) had almost the same fluency as those written by people (human-written). The participants tended to randomly select human-written news when asked to choose between two options: an excerpt from machine-generated news, and one from human-written news. Then we choose 24 of the 80 participants to evaluate the bias in the machine-generated news articles. They were able identify a bias 92\% of the times, and assigned a correct bias rating 62.91\% of the time.

\section{Related Work}
\label{section:related_work}

\noindent In this section, we discuss related work on political bias datasets, bias analysis, bias generation and detection in news articles.

\subsection{Political Bias Datasets}

In the works that study bias, \newcite{arapakis2016linguistic} collected a dataset of 561 news articles, each being labeled with 14 qualitative aspects along with article's subjectivity. Another dataset, the multi-perspective question answering (MPQA) corpus \cite{wiebe2005annotating}, contains 692 news articles, each with a label of its subjectivity. These two corpora were carefully developed with labels at the article and sentence levels. However, the labeling technique is costly to scale, and the corpora are not large enough ($<$ 1000 samples), so \newcite{chen2018learning} developed a corpus of 2,781 events from the AllSides website to characterize and flip bias in news headlines. The corpus contains news headlines and articles presented by a left-leaning and a right-leaning news source paired together with an unbiased summary of the event. However, the labeling is news source specific, so there is no information about the bias at the article level. Moreover, the corpus is not large enough to be used to generate news articles. Therefore, for this study, we used the ``All-The-News’’ dataset\ footnote{https://www.kaggle.com/snapcrack/all-the-news}.

\subsection{Bias Analysis}
Media bias has been under study for decades \cite{groseclose2005measure,fang2012mining,arapakis2016linguistic}, and various aspects of political bias have been studied from different perspectives. For example, \newcite{groseclose2005measure} quantified bias for a sample of 20 news sources in the U.S. on the basis of the number of citations used by think tanks and policy groups. Their work is among the first ones to provide clear evidence of bias in media. \newcite{lin2011more} proposed categorizing bias on the basis of variables like mentions of political parties, legislators, and ideology. Another study, \cite{chen2018learning}, focused on liberal and conservative bias, and using manual annotation, found that bias indicators usually include named entities. A more recent study explored the idea with right and left bias, and experimentally showed that named entities are indeed important, and that bias is more evident in longer texts, i.e., in full length news articles, rather than in shorter texts like sentences and paragraphs \cite{chen2018learning}. We performed the same analyses to evaluate the reliability of our dataset.

\subsection{Biased Headline Generation}
Advances in natural language processing have led to rapid development of several language generation techniques. With the release of transformer based model architectures and text representations \cite{vaswani2017attention,devlin2018bert}, machines are now able to generate high quality text outputs \cite{radford2019language}, which may or may not preserve the context. To generate text that better preserves context, researchers have studied \textit{controllable text generation}, i.e., how to rewrite a text so that it has certain attributes \cite{keskar2019ctrl,zellers2019defending}. Several of these studies demonstrated that the text style can be transferred by simply changing the relevant words in an unsupervised manner \cite{li2018delete,adelani2020generating,shen2017style}. \newcite{chen2018learning} demonstrated bias flipping in text, but only for the headlines of a news articles. To the best of our knowledge, ours is the first study on generating full-length biased news articles.

\subsection{Identification of Bias in News Articles}
\label{section:eval}

There have been several attempts in the past to identify bias as left or right at the article level \cite{zhaodeepnews,baly2018predicting,wang}, and at the source level \cite{ribeiro2018media,kulshrestha2017quantifying,an2012visualizing}. The classification of a media source as left leaning or right leaning is flawed if one starts to look at each article to identify its bias. We are more interested in the text and style of bias in news articles, and therefore, we focused on bias at the article level. At article level, \newcite{zhaodeepnews,baly2018predicting} used a smaller dataset and shallow models to classify bias at an article level using three labels. Using recent advancements in the field of natural language processing, \citet{wang} created a state-of-the-art regression model to quantify bias in news articles by using RoBERTa-based model \cite{DBLP:journals/corr/abs-1907-11692} and trained it on several datasets like the AdfontesMedia's list of articles and webhose.io\footnote{http://webhose.io/}, and so on for generalizability. We used the RoBERTa-based model to generate automatic bias ratings and evaluate bias in generated text.

\section{Dataset and Discriminativeness Ratio}
\label{section:bias_analysis}

\subsection{All The News Dataset and Automatic Bias Ratings}
The dataset we used is a collection of 139,668 full length news articles curated using the Internet Archive\footnote{https://archive.org} from 15 major news sources in the U.S. and is available on the Kaggle website under the name of ``All the news'' data\footnote{https://www.kaggle.com/snapcrack/all-the-news}. For each source, the Internet Archive was used to grab the past year-and-a-half of either homepage headlines or RSS feeds and their links were parsed through a scraper. The data obtained were not the product of scraping an entire site, but rather of scraping the more prominently placed articles. For example, CNN's articles from 5 June 2016 were what appeared on the homepage of CNN at the time of data collection. Similarly, Vox's articles from that time were everything that appeared in the Vox RSS reader, and so on. Therefore, we had a news article with its headline, publication source, publication date, and full-length body.

The collection of news articles did not have its bias ratings at the article level. We used a RoBERTa-based regression model made available to us upon requesting to ``The Bipartisan Press''\footnote{https://www.thebipartisanpress.com/political-bias-api-and-integrations/} to create bias ratings. ``The Bipartisan Press’’ annotated the data using Adfontes Media's methodology \cite{otero2019display}, which involves an initial screening and training to hire experts to annotate news articles with their bias on a scale of -42 to +42. A negative sign indicates a left-leaning bias and a positive sign indicates a right-leaning bias. We used the regression model to calculate the bias in each news article and treated these bias ratings as the ground truth. We further used the same model to evaluate the bias of the generated news articles.  Table \ref{tab:stats} lists some statistics about the ``All the news’’ dataset.
 
\begin{table}[htbp]
     \centering
     \begin{tabular}{m{0.6\linewidth}|m{0.3\linewidth}}
     \hline
          Number of news articles & 139,668  \\
          Number of unique news sources & 15 \\
          Average number of sentences in each news article & 49 \\
          Number of left biased news articles & 90,664 \\
          Number of right biased news articles & 49,004 \\
    \hline
     \end{tabular}
     \caption{``All the news'' dataset statistics.}
     \label{tab:stats}
 \end{table}

\subsection{Discriminativeness Ratio}
\label{sec:dratio} 
Bias can be found in a text if it expresses sentiment towards a specific entity ( a person,  a place, or a policy). \newcite{chen2018learning} proposed a \textit{discriminativeness ratio} to capture the fundamental difference between biased and sentimental text based on word frequency. The ratio is given as:

\begin{displaymath}
   \frac {occ(w, D_{t})}{occ(w, D_{t^{'}}) } 
\end{displaymath}
where $occ(w, D)$ is the frequency of $w$ in text $D$ and $t$ and $t^{'}$ are types of text. In biased text, $t$ and $t^{'}$ correspond to \textit{right} and \textit{left}, while in sentimental text they represent \textit{positive} and \textit{negative} sentiments, respectively. Usage of the discriminativeness ratio results in type-unrelated words having values close to 1, as they appear almost equally in both types of text. On the other hand, words that appear often in one type and rarely in the other will have higher (type $t$) and lower values (type $t^{'}$) values, respectively. 

 \begin{table}[htbp]
\centering
\begin{tabular}{m{0.25\linewidth} | m{0.20\linewidth} | m{0.25\linewidth} | m{0.20\linewidth}  }
\hline \cline{1-4}
\multicolumn{2}{l}{\textbf{Sentimental Text}}  & \multicolumn{2}{l}{\textbf{Biased Text}} \\ \hline
     \textbf{ Word }   &   \textbf{ Ratio}      &  \textbf{ Word }       &  \textbf{ Ratio}       \\ \hline
       excellent   &    220.22      &      The Atlantic        &    73.0      \\ 
        gem  &  183.99        &        Aleppo       &    64.5      \\ 
       wonderful   &    183.66          &     Ivanka Trump      &  61.0         \\ 
       \hline \hline
       mushrooms   &    1.01      &           aired     &   1.0       \\ 
       breadsticks   &    1.01      &         suspicion      &   1.0       \\ 
       dresser   &     0.99     &          recuse     &   1.0       \\ 
       \hline \hline
       unfortunately   &    $<$0.01      &          Trump      &    $<$0.01       \\ 
        terrible  &     $<$0.01     &         Truther - Breitbart       &   $<$0.01        \\ 
       rude   &    $<$0.01      &         Netanyahu       &     $<$0.01      \\ 
    \hline
\end{tabular}
\caption{Three words with highest and lowest discriminativeness ratio, and words with ratio very close to one.}
\label{table:discrim}
\end{table}

Table \ref{table:discrim} lists the words having the highest and the words having the lowest discriminativeness ratio for sentimental text and biased text. We show the results for sentimental text to simplify the explanation. The top three words in the sentimental text are positive, the bottom three are negative, and sentiment-unrelated words have a value close to one. In the biased text, the three type-unrelated words (ratio of 1.0) included both positive (“aired” and “recused”) and negative (“suspicion”) sentiment words. This is because both left- and right- biased texts use sentiment words to support and oppose entities. In addition, the top three and the bottom three biased-text words are named entities, indicating that articles with either bias tend to criticize or support different named entities, using the same words to convey sentiments. In line with this, a bias analysis by \newcite{yano2010shedding} revealed that named entities are often bias indicators.




\section{Biased News Generation}
\label{section:generation}
\begin{figure}[!ht]
    \centering
\includegraphics[width=0.9\linewidth]{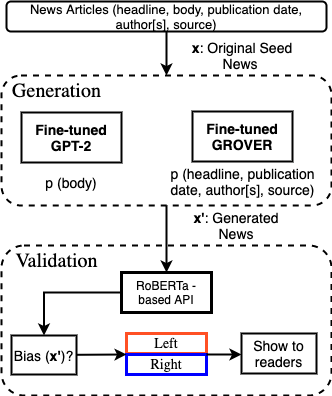}
    \caption{Biased News Generation Procedure.}
    \label{fig:gen}
\end{figure}

The most important parts of the proposed method for generating biased news is the GPT-2 text generation model \cite{radford2019language} and the controllable text generation model \cite{zellers2019defending}. As shown in Figure \ref{fig:gen}, we used a two step approach to generate biased news: generation and validation. In the generation step, an attacker provides an original news article \textbf{x} as the seed input to a generation models. The models then generate a modified article \textbf{x'} based on x. In the validation step, the generated articles are classified on the basis of bias. The attacker is assumed to have access to such a classifier and uses it to segregate left- and right- biased news. The details of our proposed method are discussed below.

\subsection{GPT-2 Model}
\label{section:gpt_generation}
The task of a language model is to learn the probability distribution of a text corpus to enable the next word to be predicted on the basis of contextual words. Given a sequence of words, $w = (w_{1}, w_{2},...,w_{T})$, the probability of the sequence is given as:

\begin{equation}
\label{eq:gpt}
     P(w) = \displaystyle \prod_{i=1}^{T} P(w_{t} | w_{1}, w_{2}, ..., w_{t-1})  
\end{equation}

Probability P(w) is calculated by learning the conditional probability of each word given a fixed number of k-context words. Many neural network architectures have been used to estimate P(w) including a feed-forward neural network \cite{bengio2003neural}, a recurrent neural network \cite{mikolov2010recurrent,sundermeyer2012lstm}, and the transformer architectures \cite{radford2018improving}. A GPT-2 model \cite{radford2019language} based on a transformer architecture has been shown to have a lower perplexity for language modeling datasets, and to generate high quality fluent texts. Therefore, we used a GPT-2 model and fine tuned it on a dataset of left- and right- biased news.

In the fine-tuning, the model was first initialized using pre-trained weights instead of random initial weights. Fine-tuning is faster than training an LM with a large number of parameters from scratch. It has been shown that fine-tuning using labeled data after initializing the model with pre-trained parameters improves the accuracy of downstream tasks \cite{devlin2018bert}. Therefore, we fine tuned the GPT-2 LM using left- biased and right- biased news.

\begin{table*}[!ht]
    \centering
    
    \begin{tabular}{m{0.96\linewidth}}
    \hline
         \textbf{Right Biased}  \\
    \hline
         \textbf{Seed Text}: President Donald Trump said he would seek to keep his tough immigration enforcement policies from harming the U.  S.  farm industry and its largely immigrant workforce, according to farmers and officials who met with him. At a roundtable on farm labor at the White House last month, Trump said he did not want to create labor problems for farmers and would look into improving a program that brings in temporary agricultural workers on legal visas.  \\
         \textit{President Donald Trump said he would seek to keep his tough immigration enforcement policies from harming the U.  S.  economy, and that the 25 countries with “green cards” had jobs to farm industry and its largely immigrant workforce according to farmers and officials who met with him last week in San Jose. \textbf{(Bias: 5.0)}} \\
    \hline
     \hline
         \textbf{Left Biased}  \\
    \hline
         \textbf{Seed Text}: Hillary Clinton and her team got in on the challenge, too, offering a special message to voters to “Don’t Stand Still, Vote Today.  ”. Her video features cameos from former President Bill Clinton along with Huma Abedin and Jon Bon Jovi. The original video had no soundtrack, so Slim Jxmmi of Rae Sremmurd added in their track “Black Beatles” to the Clinton version. Pretty hip.  \\
         \textit{Hillary Clinton and her team got in on the challenge too offering a special message to voters to Dont Stand Still Vote Today and Trump’s campaign just couldn’t do anything better.  Her video features cameos from former President Bill Clinton along with Huma Abedin and Jon Bon Jovi.  The clip was posted on YouTube in reaction to Trump’s comments, which BuzzFeed News reported. \textbf{(Bias: -13.0)}} \\
         \hline
    \end{tabular}
    \caption{Example biased news generated using fine-tuned GPT-2 LM. For the sake of brevity, only the first three sentences of original and generated  articles are presented \cite{grusky2018newsroom}. Generated text is shown in \textit{italics}.}
    \label{tab:gpt}
\end{table*}

\begin{table*}[!ht]
    \centering
    
    \begin{tabular}{m{0.96\linewidth}}
    \hline
         \textbf{Right Biased}  \\
    \hline
          \textbf{Headline}: Shaun King is Really Mad About President Trump  \\
         \textbf{Domain}: Breitbart , \textbf{Date}: June 01, 2017 , \textbf{Author[s]}: Jack Montgomery \\
         \textit{You know what is really sad? Yet another hate-filled, blame-the-victim tweet from the self-identified Brooklyn blogger and occasional Obama apologist. Wednesday night Shaun King got really fired up on social media, arguing that President Trump “took out” Chance the Rapper on Twitter — “another white, wealthy liberal take, no questions asked.” At the behest of an enraged social media following he promptly added, in his screen name, “How dare you christian white men call me racist?” \textbf{(Bias: 14.21)} } \\
    \hline
     \hline
         \textbf{Left Biased}  \\
    \hline
         \textbf{Headline}: Trump ditches press pool to play golf  \\
         \textbf{Domain}: CNN , \textbf{Date}: December 31, 2016 , \textbf{Author[s]}: Eugene Scott \\
         \textit{(CNN) — President-elect Donald Trump rode a golf cart through the course at his golf course in New Jersey on Saturday before visiting New York City to watch his son Eric Donald Trump give a New Year’s Day address. The trip marked the first time Trump has left his Trump Tower residence since he won the November election. Since the election, Trump has visited his golf courses at least once a week. He played golf Friday in New Jersey and Florida and last week in Bedminster, New Jersey. \textbf{(Bias: -11.01)}} \\
         \hline
    \end{tabular}
    \caption{Example biased news generated using fine-tuned GROVER LM. For the sake of brevity, only the first three sentences of original and generated  articles are presented \cite{grusky2018newsroom}. Generated text is shown in \textit{italics}.}
    \label{tab:grover_gen}
\end{table*}

Using techniques from \newcite{zhang2015character}, we divided the news articles from each set into training and test sets. We used a reliable implementation of the GPT-2 model available on Github\footnote{https://github.com/huggingface/transformers} to fine-tune the pre-trained model on the ``All the news’’ dataset. We used the default values for all hyperparameters. The number of training samples for left- and right- biased media were unbalanced, but since we trained a separate model for each, we had enough data for fine-tuning two good models. We fine tuned two 117M GPT-2 models, one for each type of bias. We used 85,664 and 44,004 news articles, respectively, to train the two models and 5000 each to test them for perplexity. The perplexity on the test set for the two models trained was 17.43 and 18.30, respectively, which is quite good (i.e., value less than 20 is what we look for \cite{radford2019language}). 

Finally, we generated 5000 samples for each bias type. We loaded the corresponding model and used prompts from the original articles to generate biased ones. Table \ref{tab:gpt} shows a sample for each type of bias. The generated articles are fluent and meaningful. The generated news is ``fake'' and reports incorrect factual information. For example, in the first sample, the original news has entities like \{U.S. farm industry, White House\} while the generated one completely changed them to \{U.S. Economy, San Jose\}.

 \begin{figure*}[!b]
    \centering
    \subfigure[Bias Distribution in Human Written News]
    {
        \includegraphics[width=0.46\linewidth]{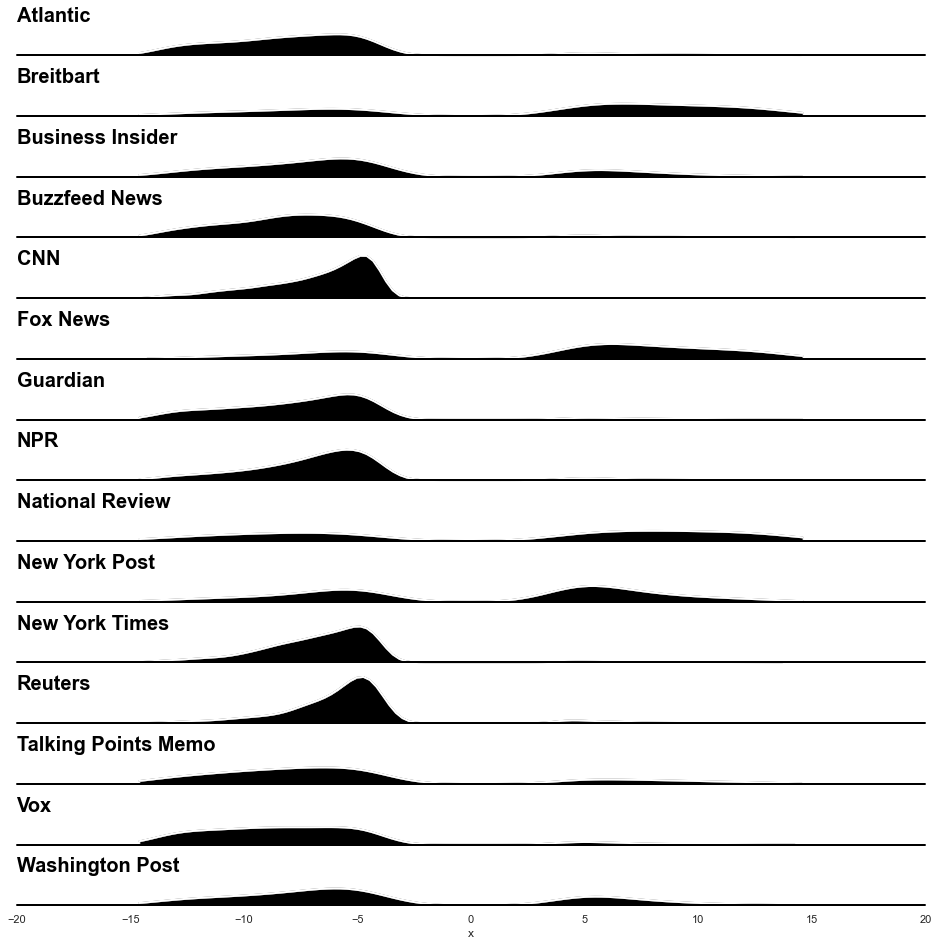}
        \label{fig:orig}
    }
    \subfigure[Bias Distribution in Machine Generated News]
    {
        \includegraphics[width=0.46\linewidth]{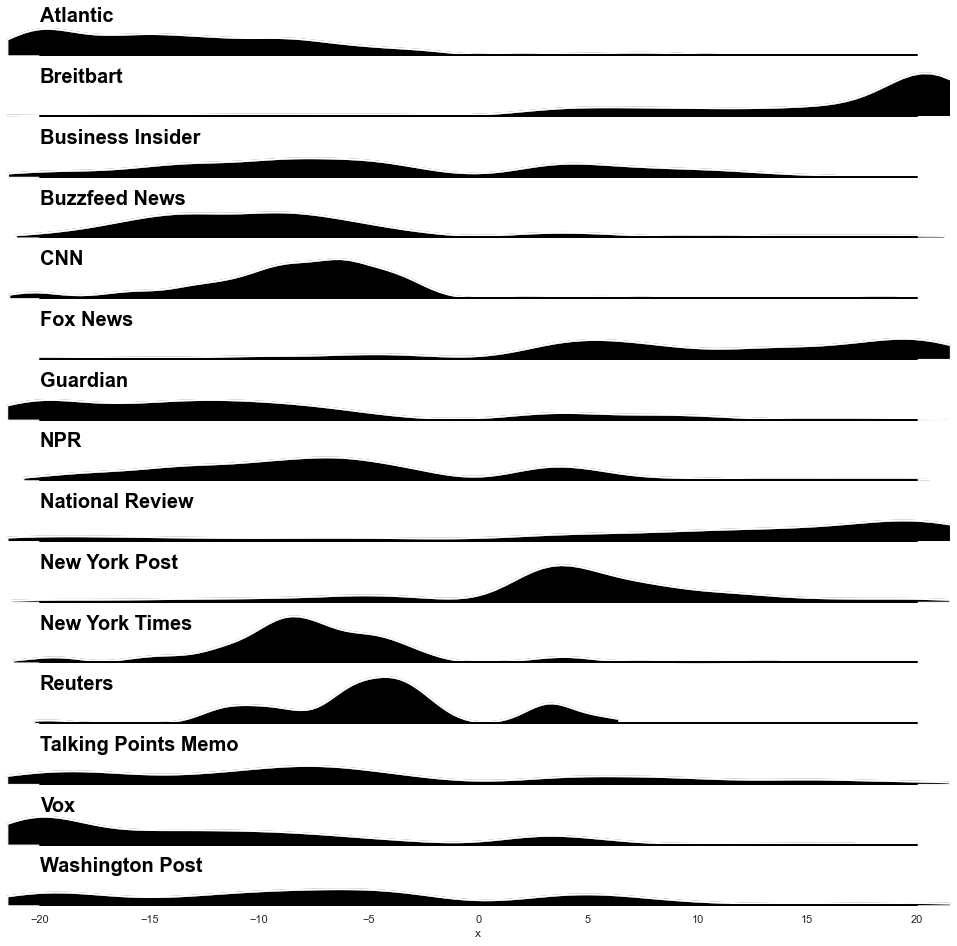}
        \label{fig:bias}
    }

    \caption{Difference in bias ratings between human-written and machine generated news (using human-written news as seed for each generation). The machine-generated news is more extreme (biased) due to being generated by fine tuned models.}
    \label{fig:bias_pub_wise}
\end{figure*}

\subsection{GROVER model}
\label{grover_generation}

The news articles generated by the GPT-2 model contain unstructured text, beginning with a $\langle start \rangle$ token and ending with an $\langle end \rangle$ token. The $\langle end \rangle$ token is particularly important as it indicates when to stop generating. However, in addition to unstructured running text, i.e., the body text, a news article has additional elements, including the publication domain, the publication date,the authors, and the headline. Generating a realistic and controlled news article requires producing all of these components. Therefore, a news article can be modeled as a joint distribution:

\begin{equation}
\label{eq:grover}
     P(domain, date, authors, headline, body)  
\end{equation}

\newcite{zellers2019defending} used the language modeling framework from equation \ref{eq:gpt} in a way that enables flexible decomposition of equation \ref{eq:grover}. GROVER starts with a set of fields $\mathbb{F}$ as context, with each field containing specific start and end tokens. To generate a target field $\tau$, we append the field specific $\langle start-\tau \rangle$ to the given context tokens to sample from the model until the $\langle end-\tau \rangle$ token is reached. For biased news generation, we fix the body of the article as the target field $\tau$ and use the other fields (domain, date, authors, headline) as context. We load pre-trained model weights to fine tunethe GROVER LM to generate biased news. 

We used the same training-test distribution as for the GPT-2 model. We defined context $\mathbb{F}$ as the set $\{ headline, date, author[s], domain \}$ and target $\tau$ as the body of the article to be generated using $F$ as context. Note that, GROVER does not need seed phrases for generation. Instead, it uses headline, date, author[s], and domain for generating the body. Table \ref{tab:grover_gen} shows a sample for each type of bias. The generated articles are fluent and appear consistent as they are presented with a domain, date, headline and author[s] names.

Figure \ref{fig:bias_pub_wise} shows the bias distributions for all the 5000 generated articles, reflecting the bias of each source. As can be clearly seen, the distributions are shifted towards the extremes for both the left- and right- biased samples, shown by the bumps being closer to the left extreme (-20) or the right extreme (+20).

\subsection{Subjective Evaluation}
To subjectively evaluate our proposed method, we asked a pool of native and non-native English speakers (annotators) to evaluate the generated biased news articles on the basis of fluency and the bias of the text. We explicitly instructed them to ignore factuality because we wanted to evaluate and validate the quality and bias of the generated articles, not their correctness. 

For evaluating quality, we considered two categories of articles: human-written ones from news sources, and machine-generated ones produced by the GPT-2 or GROVER models. The participants were asked to identify whether an excerpt was taken from a human-written, or a machine-generated article. They were shown two options to choose from, one from each class, human-written and machine-generated. Each annotator was shown ten pairs of excerpts (one human-written and one machine-generated) and asked to identify, which was the human-written one. The average selection rate was used as the metric. Further, to facilitate the evaluation, the excerpts were shortened to only three or four sentences. The evaluations were performed on a web interface with the two types of excerpts chosen randomly from two pools of samples. 

Of the 80 participants, 32 were native speakers and the rest 48 were non-native speakers. As shown in Table \ref{tab:subj_val}, the non-native English speakers tended to mark the machine-generated excerpts as human-written ones. Since the outputs from the GPT-2 and GROVER models were very similar, the ratio of participants who failed to identify the human-written news correctly was about the same for the GPT-2- and GROVER- generated samples. The lowest ratio (43\%) was for native speakers and the GROVER samples, and the highest ratio (50\%) was for non-native speakers and the GPT-2 samples. Most of the values are closer to 0.50, which indicates that the participants tended to make a random selection among the two categories of articles.

\begin{table}[htbp]
    \centering
    \begin{tabular}{c|c|c|c}
    \hline 
         \textbf{Model} &  \textbf{Native} & \textbf{Non Native} & \textbf{Overall}  \\
    \hline
          GPT-2 & 0.46 (16)  & 0.50 (23) & 0.49 (39) \\
          GROVER & 0.43 (16) & 0.48 (25) & 0.46 (41) \\
    \hline
    \end{tabular}
    \caption{Ratio of number of participants who marked machine-generated excerpt as human-written. Number of participants is shown in parentheses.}
    \label{tab:subj_val}
\end{table}

For evaluating bias, we selected 24 of the 80 participants, each having at least a college degree or who were enrolled in college at the time of annotation. We trained them to understand the media bias using various resources\footnote{https://www.coursera.org/learn/media}. Since the training was not rigorous, we made the problem simpler by treating bias as a binary variable having two values, i.e., \textit{left} and \textit{right}. For cases in which the participant was not sure, we asked them to mark the question with \textit{can't say}. Each participant was shown ten excerpts at random from the generated text and they were asked to mark their bias rating. As in the quality evaluation, only three or four sentences were shown for the sake of simplicity.  

The participants were able to identify a clear bias 92\% of the times. They marked the option of \textit{can't say} only 8\% of the time. To determine the percentage of times the participants were able to identify the bias correctly, we needed to define ``correctly’’, which is subjective. We judged that a bias rating was correct if the participant’s choice (left or right) matched that of the automatic bias evaluation . We used the API built on a RoBERTa-based model to automatically generate bias ratings for the sample excerpts shown to the participants. We found that the participants were able to identify the bias correctly 63\% of the time. The percentage might have been higher with more training and a better understanding of bias.

\section{Discussion}
Our use of the API made available to us by ``The Bipartsan Press'' to evaluate bias is a major limitation of this study. Evaluating text for bias is a very complex problem. The API was built on a RoBERTa based model trained on a dataset curated by Adfontes Media. The dataset was annotated by 20 expert annotators with at least a college degree after an extensive screening and training process\footnote{https://www.adfontesmedia.com/how-ad-fontes-ranks-news-sources/?v=402f03a963ba}. Hiring and training such annotators is expensive, and relying on non-expert annotators to calculate media bias in generated news is not promising. Since our findings conforms to the results reported by relevant literature on media bias, it is safe to assume that the results obtained using the RoBERTa-based model (with a 4\% error rate) are reliable in terms of segregating left-biased media from right-biased media.

\section{Conclusion and Future Work}
\label{section:conclusion}
\noindent We have presented a threat model and discussed how news aggregators (attackers) can manipulate readers’ opinions by flipping or increasing their bias. We described two language models generating biased news: the high-performance GPT-2 LM and the GROVER LM for controllable text generation. We used a large news article dataset to fine tune them. We used a RoBERTa-based regression model to create automatic bias ratings and to evaluate bias in generated news. Subjective evaluation of generated news articles by 80 participants suggests that they made random selections between the machine-generated and human-written news excerpts, indicating that the machine-generated news is fluent and looks similar to human-written news. Out of the 80 participants, 24 were chosen for a bias evaluation. The participants were able to see a clear bias most of the times, and marked correct bias 63\% of the times. 

For future work, techniques for a more granular control on text generation can be explored, where one can adversarially inject bias to generate twisted versions of news stories. Techniques to introduce bias during machine translation of a news article from one language to another can be explored and evaluated by comparing the generated news after translation with the news generated by non-native speakers while converting news from other languages. Apart from named entities and sentence length, there are more intrinsic patterns representing presence of bias in text. Exploration studies to find such patterns can also be done in future to better understand bias distribution in text. Another future direction can be to quantify the impact of delivering biased news to real-world users using some social media platform.


\section*{Acknowledgments}
This research was partly supported by JST CREST Grant JPMJCR18A6 and JSPS KAKENHI Grant JP16H06302 and JP18H04120, Japan.




\bibliography{naaclhlt2019}
\bibliographystyle{acl_natbib}


\appendix

\section{Supplemental Material}
\label{sec:supplemental}

\begin{table*}[!b]
    \centering
    \begin{tabular}{| m{0.20\linewidth}|m{0.20\linewidth}|m{0.20\linewidth}|m{0.20\linewidth}| }
    \hline 
         \textbf{Detector} &  \textbf{GPT-2 Generated} & \textbf{GROVER Generated} & \textbf{Overall}  \\
    \hline
          GLTR (A)  & 0.37 & 0.43 & 0.41 \\ \hline
          GPT-2 PD (B)  & 0.22  & 0.33 & 0.29  \\ \hline
          GROVER (C)  & 0.35  & 0.28 & 0.32 \\ \hline
          A + B  & 0.21 & 0.38 & 0.30 \\ \hline
          A + C  & 0.30 & 0.24 & 0.27  \\ \hline
          B + C  & 0.30 & 0.31 &  0.30 \\ \hline
          A + B + C & 0.21 & 0.24 &  0.23 \\ 
    \hline
    \end{tabular}
    \caption{Equal error rate in differentiating between human written and machine generated news. We have used three approaches independently as well as a combination of them. "+" indicates score fusion.}
    \label{tab:mach_eval}
\end{table*}

\subsection{Granularity Analysis}
Sometimes biased text segments can be identified just by looking into the title (i.e. only one sentence), as we go along the bias may or may not increase. Intuitively, as we increase the length of text tested for presence of bias, the bias should also increase. 

\begin{figure}[htbp]
    \centering
    \includegraphics[width=\linewidth]{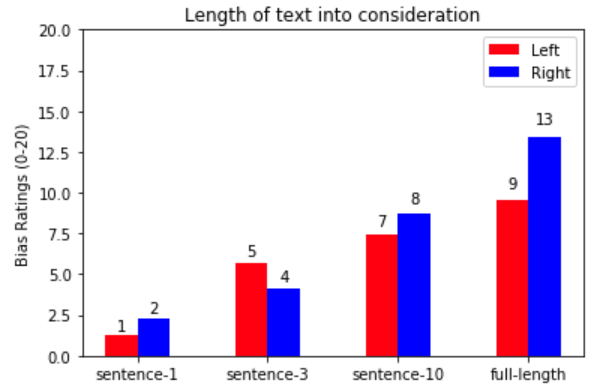}
    \caption{Granularity Analysis. The bias ratings increase as we increase the length of text to test for bias infestation.}
    \label{fig:ga}
\end{figure}

We have taken equal number of samples, i.e. 5,000, from both sides of bias. To test this hypothesis, we divided the news into 4 parts: sentence-1, which is just the title; sentence-3, first three sentences of news article \cite{grusky2018newsroom}(Lede-3); sentence-10, first 10 sentences of the news article \cite{chen2018learning}; and finally full-length, which represents the complete news. Figure \ref{fig:ga} shows that bias ratings increase as we increase the length of news being tested for bias.

\subsection{Automatic Detection}
We evaluated three automatic detection models, GLTR \cite{gehrmann2019gltr}, GROVER \cite{zellers2019defending}, and GPT-2 PD \cite{solaiman2019release} using 80 samples (news excerpts) each from human written, GPT-2 generated, and GROVER generated news. GLTR gives different probabilities of words being in top10, top100, and so on, and the other models give a probability score. We have used regression models as fusion functions while predicting with combined models. Table \ref{tab:mach_eval} shows detection results.

\end{document}